# A Multi-Task, Multi-Discipline Benchmark for Evaluating Large Language Models on the Islamic Scholarly Tradition (*turāth*)

Shahd Gaben[1], Heba Sbahi[1], Samer Rashwani[1], Abdessalam Bouchekif[1], Mutaz Al-Khatib[1], Emad Mohamed[2], Somaya Eltanbouly[1], and Mohammed Ghaly[1]

1 Hamad Bin Khalifa University, Doha, Qatar.

2 Nazarbayev University, Astana, Kazakhstan.

## Abstract

Large language models (LLMs) are increasingly used for question answering, education, and research, including in religious and cultural domains where answers depend on specialised source traditions. Yet in Islamic Studies, key concepts, methods, and debates preserved in the authoritative scholarly tradition, known as *turāth*, lack high-quality annotated resources. We introduce IslamicTurathBench (ISTB), a multi-task, multi-discipline dataset for evaluating LLMs on classical Islamic scholarship. Developed and reviewed by domain experts, ISTB contains 3,465 question–answer items drawn from 35 recognised source works spanning over 12 centuries of scholarship across seven key fields of Islamic Studies. To enable comprehensive profiling of model capabilities, ISTB is structured along two axes: scholarly demand (Beginner, Intermediate, and Advanced) and task format (multiple-choice questions, passage-based comprehension, and open-ended knowledge questions). ISTB includes aggregated scores from a scholarly human reference panel and zero-shot baselines from ten systems. The dataset supports reproducible evaluation of language-model behaviour across source works, disciplines, scholarly demand levels, and question formats in a historically layered scholarly domain.

## 1 Background & Summary

Large language models (LLMs) are increasingly deployed as assistants for question answering, education, research support, and public information access.[1–3] Their utility in these settings depends not only on fluent language generation but also on whether their outputs reflect the authoritative sources, concepts, and standards of the target domain. This is particularly acute in fields where knowledge is transmitted through specialised texts and well-established scholarly methods. In such domains, superficially plausible responses are often incomplete, poorly grounded, or inconsistent with how experts would address the subject.

Religious and cultural knowledge domains exemplify this challenge. They frequently involve inherited source traditions, discipline-specific terminology, and interpretive conventions that cannot be captured by general language fluency alone. Islamic Studies is a prime example; the vast corpus of its concepts, debates, and methods is preserved in the classical Islamic scholarly tradition, commonly known as *turāth*. This tradition encompasses foundational works on *Quran* Sciences (*ʿUlūm al-Qurʾān*), Hadith Sciences (*ʿUlūm al-Ḥadīth*), Theology (*ʿAqīda*), Principles of Jurisprudence (*Uṣūl al-Fiqh*), Jurisprudence (*Fiqh*), Sufism (*Taṣawwuf*), and Prophetic Biography (*Sīra Nabawiyya*). These fields are intellectually interconnected but distinct; each has its own core texts, levels of pedagogical progression, technical vocabulary, and epistemic criteria for what counts as an adequate answer and reliable knowledge.

Evaluating language models in this setting, therefore, requires methodologies that extend beyond assessing simple Arabic text generation or general religious questions. A robust benchmark must systematically evaluate whether a model can handle diverse Islamic disciplines, graded levels of scholarly demand, and varied cognitive task formats. It should also preserve granular metadata regarding the source works from which questions were developed, thereby enabling researchers to analyse model performance in relation to the specific scholarly material being tested.

Existing Arabic and Islamic language resources provide important foundations for this work. General Arabic datasets support tasks such as question answering, reading comprehension, text classification, and instruction following. Islamic-domain datasets cover valuable areas such as *Quran* question answering, Hadith retrieval, fatwa classification, and legal or ethical question answering. These resources have advanced Arabic and Islamic natural language processing, but they do not fully span the evaluation space needed for classical Islamic scholarship. Many focus on a single source type, a narrow subdomain, contemporary web-facing discourse, or a solitary question format. As a result, they do not offer a comprehensive framework to compare model behaviour across classical source works, multiple Islamic disciplines, graded scholarly demand levels, and diverse evaluation formats. Specifically, the operational mismatch between web-centric benchmarks and the rigorous demands of classical text traditions leaves a critical evaluation gap.

We introduce IslamicTurathBench (ISTB), a multi-task, multi-discipline Arabic benchmark dataset specifically designed to bridge this evaluation gap by facilitating rigorous diagnostics on classical Islamic scholarship. The dataset contains 3,465 expert-written question–answer items drawn from 35 recognised source works across seven fields of Islamic Studies: *Quran* sciences, Hadith sciences, Islamic theology, jurisprudence, principles of jurisprudence, Sufism, and Prophetic biography. Each item is enriched with granular metadata for its source work, discipline, scholarly demand level, and task format, enabling users to filter the dataset or report results at varying levels of granularity.

Table 1 positions IslamicTurathBench (ISTB) against related resource types and benchmarks along the dimensions most relevant to this identified gap: source base, disciplinary breadth, and task format. The table aims to show complementarity rather than replacement. Existing resources remain valuable for their target settings, while ISTB contributes a dataset uniquely focused on the classical Islamic scholarly tradition across multiple disciplines and task formats.

ISTB is structured along two main axes. The first axis is scholarly demand, represented by three scholarly demand levels: Beginner, Intermediate, and Advanced. These levels reflect progression from introductory source works to more specialised works requiring greater scholarly depth. The second axis is task format. The dataset includes multiple-choice questions, passage-based comprehension questions, and open-ended knowledge questions, where the model receives neither a source passage nor answer options. Together, these axes form a 3 × 3 evaluation design that supports rigorous comparisons across both the level of knowledge being tested and the kind of answer expected.

The dataset was developed and reviewed by domain experts. Its construction combined expert selection of source works, manual question authoring, task-specific answer design, structured review, and a subsequent refinement audit. The released dataset is provided in a row-oriented format with separate JSON and CSV files for each task type, together with source metadata, schema documentation, validation checks, and loading code. Aggregated scores from a 148-item scholarly human reference panel are included as a reference layer, while raw human response files are excluded to protect participant privacy. Furthermore, we provide empirical usage illustrations based on zero-shot evaluations of ten language-model systems, including frontier, open-weight, Arabic-specialised, and Islamic-domain systems.

*Table 1. Peer Islamic NLP benchmarks compared along structural dimensions relevant to scholarly evaluation.*

| Benchmark | Source layer | Disciplinary scope | Scholarly demand rubric | Task format |
|---|---|---|---|---|
| **IslamicLegalBench**[4] | 38 foundational Islamic jurisprudence texts across seven schools | Islamic jurisprudence | Low / Moderate / High complexity | 13 legal-reasoning task types; judged model responses |
| **IslamicMMLU**[5] | Quran, Hadith, and Fiqh question banks | Quran, Hadith, Islamic jurisprudence | No explicit scholarly demand rubric | MCQ |
| **QIAS 2025 — Inheritance**[6] | IslamWeb fatwas and inheritance case resolutions | Islamic inheritance | Beginner / Intermediate / Advanced | 6-option MCQ |
| **QIAS 2025 — Islamic Assessment**[6] | 25 traditional Islamic reference works | Classical Islamic knowledge across specialised disciplines | Beginner / Intermediate / Advanced | 4-option MCQ |
| **MAWARITH**[7] | Generated Arabic inheritance cases enriched with expert-reviewed reasoning and juristic justifications | Islamic inheritance | No explicit scholarly demand rubric | Structured reasoning-chain answers |
| **FiqhQA**[8] | Islamic rulings categorised by four Sunni schools | Islamic jurisprudence | No explicit scholarly demand rubric | Open-ended ruling generation |
| **IslamicFaithQA**[9] | Bilingual Islamic QA items with atomic single-gold answers | Islamic question answering | No explicit scholarly demand rubric | Open-ended generative QA |
| **IslamicTurathBench** | 35 recognised works from the classical Islamic scholarly tradition | Seven Islamic studies fields | Beginner / Intermediate / Advanced | MCQ, passage-based open-ended comprehension, open-ended QA |

ISTB offers several key contributions to the field:

- It provides a source-grounded Arabic benchmark dataset built principally from recognised works in the classical Islamic scholarly tradition, rather than from undifferentiated web material or contemporary advice discourse.
- It covers seven fields of Islamic Studies, enabling evaluation beyond a single scripture-centred, legal, or advice-oriented subdomain.
- It combines three levels of scholarly demand with three task formats, offering researchers a systematic framework to compare model behaviour across scholarly demand levels and answer conditions.
- It includes expert review, validation artefacts, aggregated human reference scores, and model baseline results to support reproducible evaluation and future comparison.

Together, these features make ISTB a highly reusable dataset for researchers in language technology, Arabic natural language processing, digital humanities, religious studies, and Islamic Studies. It supports the systematic evaluation of language-model behaviour in a historically layered Arabic scholarly domain while keeping the dataset structure sufficiently transparent for reuse, extension, and critical audit.

# 2 Methods

The dataset was developed by a collaborative team of domain experts in Islamic Studies to support the scientific community with two primary features: (1) a rigorous evaluation framework anchored in recognised works used in Islamic scholarly learning rather than brief web-based summaries; and (2) a metadata-rich architecture that allows researchers to conduct both broad aggregate evaluations and granular analyses by discipline, scholarly-demand level, and task format.

The final dataset contains 3,465 expert-authored question–answer items drawn from 35 source works, spanning seven Islamic Studies disciplines, three scholarly demand levels, and three task formats: multiple-choice questions (MCQ), comprehension questions based on supplied source passages (COMP), and open-ended knowledge questions (KNOW). Coverage is virtually balanced across disciplines, ranging from 491 to 497 items per discipline. The scholarly demand distribution contains 1,400 Beginner items, 1,052 Intermediate items, and 1,013 Advanced items, whilst the task-format distribution contains 2,276 MCQ items, 417 COMP items, and 772 KNOW items.

The development of the dataset progressed through five stages: defining the evaluation goals; establishing the dataset design decisions and taxonomy; defining item-writing criteria to ensure that questions were answerable, clear, source-aligned, and evaluable; constructing and drafting the dataset; and refining the corpus to reach the final released version. Figure 1 illustrates the end-to-end pipeline from initial design to the final dataset.

A human reference layer was also collected on a stratified 148-item subset to provide the community with a scholarly baseline for comparison. This panel is intended as a generalist scholarly reference baseline, not as a ceiling on specialist performance. The design and administration of this reference layer are described in the final subsection.

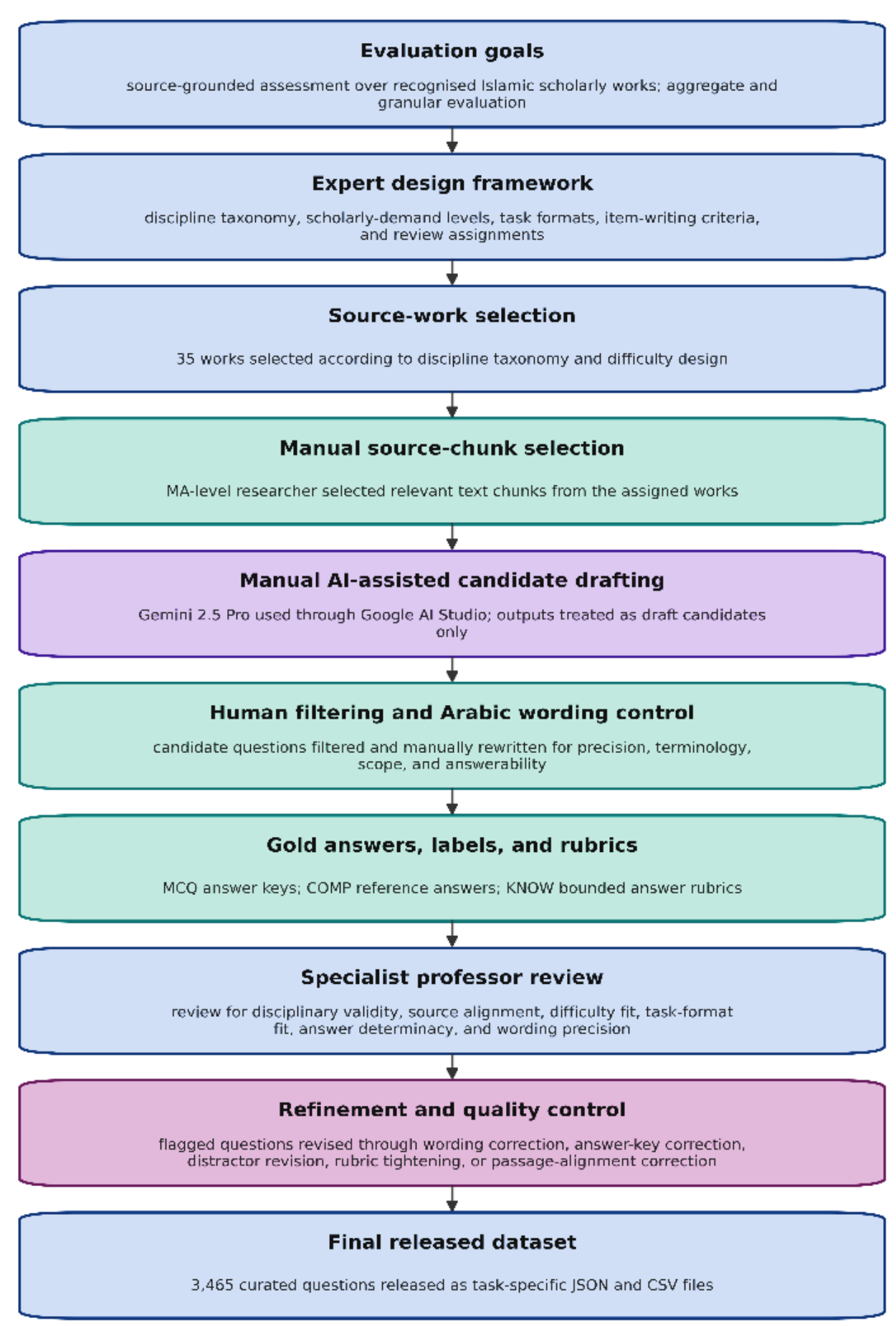

*Figure 1. IslamicTurathBench construction pipeline. The dataset was produced through expert goal setting, taxonomy and source-work design, manual source-chunk selection, AI-assisted candidate drafting, human filtering and Arabic wording control, gold-answer and rubric setting, specialist academic review, refinement, and release formatting. Automated model outputs were used solely as draft candidates or item-quality flags; final inclusion and revision decisions were made exclusively by human reviewers.*

## 2.1 Expert Authoring Team

The dataset was developed, authored, and validated by a specialised team of Islamic Studies experts, including three university faculty members with doctoral expertise across the benchmark's targeted disciplines, referred to hereafter as "the professors", and an MA-level Islamic Studies researcher, referred to hereafter as "the researcher". The professors defined the disciplinary taxonomy, selected the source texts, designed the evaluation axes, and conducted specialist reviews. The researcher oversaw source-chunk selection, candidate filtering, linguistic control, item-level verification, and dataset curation. Together, the team implemented multi-phase quality control to ensure that the final questions were answerable, source-aligned, linguistically precise, and evaluable against a reference answer. These criteria were particularly important for open-ended knowledge questions (KNOW), where legitimate scholarly plurality can make evaluation unstable unless the expected answer is explicitly bounded.

*Table 2. Scholarly disciplines and selected source works in IslamicTurathBench. The table summarises the seven Islamic scholarly disciplines covered by the benchmark, their functional role within the classical Islamic scholarly tradition (turāth), and the selected Arabic source texts used for item construction. Dates in parentheses indicate the author's death date in Hijri/Gregorian years.*

| **Scholarly Discipline** | **Role in Classical Scholarship (*Turāth*)** | **Selected Source texts (Arabic title; death date Hijri/Gregorian)** |
|---|---|---|
| **Quran Sciences (*ʿUlūm al-Qurʾān*)** | Preserving the textual transmission, contexts of revelation (*asbāb al-nuzūl*), and rules of interpretation of the literal word of God. | • فنون الأفنان في عيون علوم القرآن لابن الجوزي (597هـ/1201م)<br>• أنوار التنزيل وأسرار التأويل لناصر الدين البيضاوي (1286/685)<br>• الإتقان في علوم القرآن للسيوطي (1505/911)<br>• الزيادة والإحسان في علوم القرآن لابن عقيلة المكي (1737/1150)<br>• مباحث في علوم القرآن لمناع القطان (1999/1420) |
| **Hadith Sciences (*ʿUlūm al-Ḥadīth*)** | Systematising the transmission, authenticity (*ṣiḥḥah*), and moral authority of the statements and deeds of the Prophet. | • مشكل الآثار للإمام الطحاوي (933/321)<br>• مقدمة ابن الصلاح لابن الصلاح (1245/643)<br>• شرح صحيح مسلم للنووي (1277/676)<br>• شرح نخبة الفكر لابن الهمام (1457/861)<br>• فتح المغيث بشرح ألفية الحديث للسخاوي (1497/902)<br>• منهج النقد في علوم الحديث لنور الدين عتر (2021/1442) |
| **Theology (*ʿAqīdah*)** | Establishing the rational and scriptural foundations of belief, the nature of the Creator, and cosmic existence. | • المواقف في علم الكلام لعضد الدين الإيجي (1355/756)<br>• شرح المقاصد في علم الكلام للتفتازاني (1390/793)<br>• إتحاف المريد بشرح جوهرة التوحيد للقاني (1668/1078)<br>• العقيدة الإسلامية لحسن حبنكة الميداني (1978/1398) |
| **Principles of Jurisprudence (*Uṣūl al-Fiqh*)** | Formulating the legal theory, linguistic rules, and methodologies required to derive practical rulings from scriptural sources. | • تقويم الأدلة في أصول الفقه لأبي زيد الدبوسي (1039/430)<br>• المستصفى للغزالي (1111/505)<br>• نهاية السول للإسنوي (1370/772)<br>• تشنيف المسامع للزركشي (1392/794)<br>• علم أصول الفقه لعبد الوهاب خلاف (1956/1375) |
| **Jurisprudence (*Fiqh*)** | Mapping the detailed, practical rulings governing various aspects of life, including ritual worship (*ʿibādāt*) and social transactions (*muʿāmalāt*). | • إحياء علوم الدين للغزالي (1111/505)<br>• منهاج الطالبين وعمدة المفتين للنووي (1277/676)<br>• أنس المطالب لزكريا الأنصاري (1520/926)<br>• حاشية الرملي الكبير للرملي (1550/957)<br>• مغني المحتاج للخطيب الشربيني (1570/977)<br>• الفقه المنهجي على المذهب الشافعي لمصطفى الخن (2008/1429) ومصطفى البغا وعلي الشربجي |
| **Sufism (*Taṣawwuf*)** | Purifying the inner self (*nafs*), cultivating moral character (*akhlāq*), and fostering a deeper spiritual connection to creation. | • التعرف لمذهب أهل التصوف للكلاباذي (990/380)<br>• الرسالة القشيرية للقشيري (1072/465)<br>• الحكم العطائية لابن عطاء الله السكندري (1309/709)<br>• مدارج السالكين لابن قيم الجوزية (1350/751)<br>• مدخل إلى التصوف لأبي الوفا التفتازاني (1994/1415) |
| **Prophetic Biography (*Sīrah*)** | Providing the historical, situational, and biographical details of the Prophet's life as the living instantiation of the governing religio-moral system of Islam (Sharia). | • السيرة النبوية لابن هشام (829/213)<br>• الروض الأنف لعبد الرحمن السهيلي (1185/581)<br>• الرحيق المختوم لصفي الرحمن المباركفوري (2006/1427)<br>• فقه السيرة للبوطي (2013/1434) |

## 2.2 Discipline taxonomy and selection criteria

The source corpus was delineated using an internal discipline taxonomy rather than relying on external topical clustering or generic semantic tags. The seven disciplines were adopted from classical Islamic knowledge taxonomies, specifically drawing upon authoritative works in this domain, including Ibn al-Akfānī's (749 AH / 1348 CE) *Irshād al-Qāṣid*, Ibn Khaldūn's (808 AH / 1406 CE) *Muqaddimah*, and Ḥājjī

Khalīfah's ( 1067 AH / 1657 CE) *Kashf al-Ẓunūn*[10–12]. Overlapping historical categories were consolidated into seven contemporary scholarly fields: Quran sciences, Hadith sciences, Islamic theology, jurisprudence, principles of jurisprudence, Sufism, and Prophetic biography.

These classifications represent formal scholarly fields with established internal logics within Islamic scholarship. Together, they cover textual transmission and interpretation, prophetic reports, theology, legal theory, practical jurisprudence, spiritual formation, and prophetic biography. They also correspond to many of the doctrinal, legal, ethical, historical, and devotional questions that Muslims continue to raise and discuss in contemporary digital environments. By anchoring IslamicTurathBench in these disciplines, the dataset is both grounded in the structure of the classical Islamic scholarly tradition and relevant to present-day AI-mediated religious information seeking. Table 2 summarises the seven disciplines selected for IslamicTurathBench, their role within the *turāth*, and the source texts used to construct the benchmark. The complete source-work inventory is tabulated in the source_links.csv file within the data repository. The jurisprudence (*fiqh*) questions are strictly bound by the selected source works listed in the inventory; the dataset does not assert comprehensive coverage of all historical jurisprudential schools or subtraditions.

## 2.3 Design axes: scholarly demand × task format

ISTB was engineered along two primary axes: scholarly demand and task format. Scholarly demand indicates the depth of the disciplinary knowledge required, while task format controls the context provided at inference time and the expected response *structure*.

*Table 3. Scholarly-demand tier design in IslamicTurathBench. The table defines the three scholarly-demand tiers used in the benchmark by linking each tier to its pedagogical stage, source-text profile, core scholarly operations, and corresponding cognitive-demand categories in Bloom's revised taxonomy.*

| Scholarly-demand tier | Pedagogical Stage | Source-Text Profile | Core Scholarly Operations | Cognitive Demand (Bloom's Taxonomy) |
|---|---|---|---|---|
| **Beginner** | Introductory stage of learning | Introductory manuals, primarily written by contemporary scholars using accessible Modern Standard Arabic. | Basic conceptualisation, direct recall, and local text inference. | Remembering and Understanding |
| **Intermediate** | Intermediate stage of learning | Standard disciplinary treatises and annotated textbooks. | Validation of textual propositions, relation- and evidential tracking. | Applying and Analysing |
| **Advanced** | Advanced stage of learning | Advanced works, multi-volume commentaries, and primary authoritative sources. | Analysing the underlying reasoning process, weighing conflicting views, and deriving new case rulings. | Evaluating and Creating |

**Scholarly demand.** The scholarly demand axis is operationalised through a unified framework that fuses two dimensions in tandem: the depth of the source work (the source-text level) and the cognitive operation required by the question itself (the question level). At the source-text level, works are stratified according to the pedagogical principle of *al-tadarruj fī ṭalab al-ʿilm* – the staged progression of knowledge acquisition – as reflected in classical discussions of learning in *al-Zarnūjī's* (591 AH / 1195 CE) *Taʿlīm al-Mutaʿallim* and *Ibn Khaldūn's Muqaddimah*[10,13]. At the question level, the expected operation is specified using Bloom's revised taxonomy[14]. Under this design, each of the three scholarly demand tiers represents a harmonious intersection between classical Islamic pedagogical stages and the cognitive operations delineated in Bloom's revised taxonomy[14], systematically structured in Table 3.

**Question format.** The format axis controls inference-time context and expected response structure, mitigating the risk of collapsing varying levels of scholarly demand into a single evaluation format[15–20]. MCQ questions provide four options and require single-option selection. COMP questions supply a source passage and require an open-ended Arabic response grounded strictly in that text. KNOW questions supply neither answer options nor source passages and require an open-ended Arabic response.

Across all formats, questions were designed to be answerable against a determinate reference structure. The default design principle was that each item should have one correct answer, or one complete set of required answer elements, rather than an open-ended space of possible responses. However, some Islamic Studies questions naturally admit more than one correct answer while remaining finite and evaluable. In the released dataset, 32 KNOW questions use this controlled-plurality structure; their reference answers consist of finite lists of acceptable responses.

## 2.4 Dataset construction and quality control

The transition from the conceptual framework to the final dataset followed the expert-led construction and quality-control pipeline summarised in Figure 1. Following the framework design, the professors selected the 35-source works. The researcher then conducted a close reading of the selected works and manually extracted relevant textual chunks containing self-contained, testable scholarly concepts. These chunks served as the grounding for subsequent question drafting.

To scale drafting, the researcher used Gemini 2.5 Pro manually as a bounded drafting aid through Google AI Studio[21-23]. For each generation, the researcher supplied the manually selected text chunk, target scholarly demand tier, task format, and answer-label or rubric constraints. She then filtered the candidate drafts, discarding questions that failed to comply with the taxonomy, scholarly demand criteria, task format, or quality rubric. Retained candidates underwent manual Arabic editing to control wording, terminology, scope, and answerability. The question-drafting prompts are provided in the Supplementary Materials (Section S4).

The questions were submitted to the professors for validation according to their areas of expertise. They audited questions for disciplinary validity, source alignment, scholarly demand fit, task-format fit, and wording precision. Questions flagged during professor review or subsequent pilot evaluation runs were not automatically discarded; they were systematically revised when the issue could be resolved while preserving the dataset's structural balance. Refinement focused on tightening wording, adjusting MCQ distractors, correcting gold answers, refining grading rubrics, and ensuring passage alignment.

## 2.5 Human reference panel construction

To provide the scientific community with a baseline for comparative analysis, a human reference layer was collected using a stratified subset of 148 questions (46 MCQ, 60 COMP, 42 KNOW). The subset covered all seven disciplines, all three scholarly demand levels, and all three task formats.

The panel comprised three Islamic Studies professors with complementary expertise across the dataset's disciplines, and hence, is designed to provide a generalist scholarly baseline, not a specialist ceiling. Participants were tested across all disciplines, scholarly-demand levels, and task formats represented in the subset, reflecting the broad foundational knowledge expected of a well-rounded Islamic Studies scholar rather than restricting each participant to their narrowest sub-specialisation.

The evaluation was administered digitally through separate Google Forms for the three task formats. Participants were given a 72-hour window to complete the evaluation at their own pace, with the technical ability to review and edit responses before final submission. The protocol prohibited external consultation. Participants were instructed to rely on their own knowledge for MCQ and KNOW questions, and on the supplied passage for COMP questions, without using external references, texts, search engines, or other sources. Optional comment fields allowed participants to flag issues related to question clarity, text excerpts, answer options, or other concerns.

Participants were compensated for their time at a rate equivalent to three working days, corresponding to the three-day completion window. Compensation was not contingent on performance, score, or completion speed. The study coordinator retained identity information only for logistical communication and compensation processing. Raw human responses, individual-level records, and personally identifiable information are excluded from the public dataset release. Only aggregated, de-identified scores and the corresponding subset identifiers are released to support computational benchmarking.

# 3 Data Records

IslamicTurathBench (ISTB) v1.0 is publicly available on Zenodo under DOI 10.5281/zenodo.2067493024 and is released under a Creative Commons Attribution 4.0 International (CC BY 4.0) license. The deposit contains three task collections in JSON and CSV format, bibliographic metadata for the 35-source works, an integrity manifest with SHA-256 checksums and per-task schemas, a lightweight reference loader, build-level metadata, aggregate corpus statistics, and an aggregated scholarly human-reference layer on the 148-item subset. All files are UTF-8 encoded; CSV files use UTF-8 with byte order mark (utf-8-sig). No proprietary format is required to read the data.

## 3.1 Deposit contents and repository layout

```
islamicturathbench-v1.0-flat/
├── README.md
├── LICENSE
├── MANIFEST.json        # file inventory, per-task schemas, and SHA-256 checksums
├── source_links.csv     # bibliographic metadata for the 35 source works
├── load_benchmark.py    # data loader
├── _deposit_meta.json   # build-level metadata
├── data/
│   ├── README.md
│   ├── istb_mcq.json  / istb_mcq.csv
│   ├── istb_comp.json / istb_comp.csv
│   └── istb_know.json / istb_know.csv
├── statistics/
│   └── unified_data_stats.json
└── human_panel/
    ├── subset_question_ids.json
    └── human_reference_panel_aggregate.json
```

*Figure 2. Repository layout of the IslamicTurathBench v1.0 Zenodo deposit. The tree shows the root documentation files, task-specific JSON and CSV files, source-work metadata, integrity and schema metadata, loading code, aggregate corpus statistics, and aggregated human reference panel files.*

The package deposited is organised as shown in Figure 2. MANIFEST.json records the package version, encoding conventions, task labels, row counts, total question count, column lists, source-link columns, file sizes, and SHA-256 checksums for the inventoried release files. source_links.csv lists the 35-source works used in the benchmark using the columns: source_text (Arabic/transliterated/English), Discipline (Arabic/English), source_url, author (Arabic/English), author death_date (Hijri/Gregorian), and book citation (Arabic/English). _deposit_meta.json records build-level metadata for the deposited package.

statistics/unified_data_stats.json provides machine-readable aggregate corpus statistics derived from the released files.

## 3.2 Task collections

The benchmark comprises 3,465 expert-authored question–answer items across three task collections: multiple-choice questions (MCQ), passage-based open-ended comprehension questions (COMP), and closed-book open-ended knowledge questions (KNOW). Each task collection is released in both JSON and CSV format (Table 4).

*Table 4. Task collections in the released ISTB v1.0 package.*

| Task file stem | Task collection | Formats | Rows |
|---|---|---|---|
| **istb_mcq** | Multiple-choice questions (MCQ) | JSON + CSV | 2,276 |
| **istb_comp** | Passage-based open-ended comprehension (COMP) | JSON + CSV | 417 |
| **istb_know** | Closed-book open-ended questions (KNOW) | JSON + CSV | 772 |
| **Total** | | | 3,465 |

Each task JSON file is a top-level array of row objects. The corresponding CSV file exposes the same columns as the JSON file. Within each task file, rows are sorted by scholarly_demand_level and then by natural numeric-aware question_id.

## 3.3 Field dictionary

All task files share a common set of identifying and descriptive fields, with additional fields specific to MCQ, COMP, or KNOW items. Table 5 defines the released row-level fields.

*Table 5. Field dictionary for the released task files.*

| Field | Appears in | Type | Description |
|---|---|---|---|
| **source_text_title** | all task files | string | Arabic title of the source work from which the item is derived. |
| **scholarly_demand_level** | all task files | string | Scholarly demand tier: Beginner, Intermediate, or Advanced. |
| **discipline** | all task files | string | Arabic label of the Islamic studies discipline. |
| **question_id** | all task files | string | Globally unique item identifier. |
| **question** | all task files | string | Arabic question text. |
| **text_chunk_id** | COMP | string | Identifier of the source-passage unit supplied with the COMP question. |
| **text_chunk** | COMP | string | Arabic source passage supplied as context for the COMP question. |
| **choice_A–choice_D** | MCQ | string | Four answer options for a multiple-choice item. |
| **correct_answer** | MCQ, COMP | string | For MCQ, the correct option label (A, B, C, or D); for COMP, the Arabic reference answer. |
| **possible_item_1–possible_item_8** | KNOW | string | Accepted answer element or complete reference answer, depending on the value of no_requested_items. |
| **no_requested_items** | KNOW | integer or null | Number of acceptable answer elements required for a complete KNOW response; null indicates a single complete reference answer. |
| **answer_has_Quran_verse** | COMP, KNOW | boolean | Indicates whether the reference answer contains *Qur'anic* material. |
| **answer_has_hadith** | COMP, KNOW | boolean | Indicates whether the reference answer contains Hadith material. |

MCQ rows contain four Arabic answer options and one correct option label. COMP rows contain a source passage and a reference answer. The 417 COMP rows draw on 88 distinct passage units; rows sharing the same text_chunk_id repeat the same passage text. KNOW rows use possible_item_1 through possible_item_8 to encode bounded reference-answer rubrics. When no_requested_items is null, the item has a single complete reference answer, stored in possible_item_1; if that answer is enumerative, the complete listed answer is required. When no_requested_items is 1, any one non-empty possible_item_* entry is sufficient for a complete answer. When no_requested_items is an integer greater than 1, a response must contain at least that number of non-empty possible_item_* entries to be considered complete. Empty unused possible_item_* slots are stored as empty strings in JSON and as empty cells in CSV.

This schema keeps all released items explicitly bound by reference answers. It also allows users to filter the benchmark by source work, discipline, scholarly demand tier, and task format.

### 3.4 Human reference panel files

The human_panel/ directory contains the aggregated scholarly human reference layer. The file subset_question_ids.json identifies the 148 benchmark items used in the human reference evaluation and provides subset counts by task format, scholarly demand level, and discipline. The corresponding benchmark rows, including question text, source passages, options, and reference answers, are available in the released task files and can be retrieved by filtering on question_id.

The file human_reference_panel_aggregate.json reports aggregate panel scores derived from three anonymous Islamic Studies professors on a 0.0–1.0 scale. It includes the overall human reference score, task-format marginals, scholarly demand marginals, discipline marginals, task-by-scholarly demand means, and task-by-scholarly demand-by-discipline means. Individual evaluator identities and raw human responses are not included in the public release.

### 3.5 Loading and encoding

Users may load the benchmark through load_benchmark.py or read the JSON and CSV files directly from the data/ directory. JSON is recommended for programmatic use because it preserves booleans, null values, and Arabic strings without spreadsheet coercion. The reference loader requires Python 3.7 or later and uses only the Python standard library. When using CSV files in spreadsheet software, users should import the files explicitly as UTF-8 rather than opening them by double-clicking, to avoid corruption of Arabic text or coercion of identifier-like values.

## 4 Technical Validation

We validate three structural layers of the release: the set of source works as a reusable scholarly foundation, the question design as a bounded scoring instrument within a pluralistic scholarly tradition, and the refinement audit as evidence of the quality of the final reference answers and item labels.

## 4.1 Corpus design for evaluation and downstream reuse

IslamicTurathBench is anchored to a set of source works designed to support both standardised evaluation and downstream computational development. The corpus does not merely serve as a reservoir for question generation; it constitutes a staged reference layer spanning seven Islamic disciplines. Because each dataset item is linked to a source title—with bibliographic metadata and URLs provided in the release package—users can build retrieval, fine-tuning, corpus-analysis, or retrieval-augmented generation (RAG) systems over the documented source layer, where source texts are digitally accessible, and subsequently evaluate those systems against source-linked questions.

**Pedagogical anchoring.** The corpus design validates the benchmark against the progression used in Islamic scholarship rather than externally imposed topic clusters. The Beginner, Intermediate, and Advanced tiers operationalise a staged progression of study. Beginner items correspond to introductory texts with direct knowledge targets; Intermediate items correspond to commentaries that support comparison and evidential reasoning; and Advanced items correspond to specialised treatises requiring application and synthesis.

**Downstream utility.** While the works were selected based on internal Islamic pedagogical structures, they were simultaneously validated for their computational utility. To support end-to-end Natural Language Processing (NLP) pipelines—such as corpus pre-training, fine-tuning, or Retrieval-Augmented Generation (RAG)—the dataset prioritises fully digitised texts. With the exception of two books currently linked to scanned PDFs, all source texts in the corpus are linked to machine-readable formats, with exact access URLs provided in the accompanying source_links.csv file. This design preserves the documented corpus as a reusable technical object. Downstream users can filter evaluation items by discipline, scholarly demand, or task format; build a retrieval index over the digitally accessible works; compare closed-book and retrieval-supported systems; or use the questions as a diagnostic test following model fine-tuning. The validation claim is therefore not that IslamicTurathBench exhausts the entirety of historical Islamic scholarship, but rather that it establishes a traceable, staged, and computationally reusable foundation for targeted model evaluation.

## 4.2 Question design for tractable scoring in a plural scholarly tradition

A central technical challenge in evaluating Islamic Studies is the presence of legitimate plurality. In this context, plurality means that different recognised scholars or source works may give different but valid answers to the same broad question. Without explicit constraints, open-ended queries may admit multiple divergent but equally valid answers, rendering automated or human scoring highly unstable. The benchmark therefore enforces boundedness as a question-validity requirement.

**Format-level constraints. The** three task formats define the epistemic support available at inference time and inherently bound the acceptable answer space. While MCQ constraints are structural (four options, one key), the open-ended formats rely on explicit contextual bounds. COMP answers are tethered to the supplied source passage, while KNOW questions specify the named source, author, or school whenever necessary, and are bound by a finite accepted answer set.

**Item-level bounding.** The expert-review criteria functioned as targeted risk-control mechanisms (Table 6). For example, source fidelity requirements prevent models from generating loosely attributed answers to the "Islamic tradition" in general, while determinacy requirements prevent unmarked multiple correct answers. The goal was not to artificially erase scholarly plurality from the domain, but to delineate the scope of each question so clearly that responses could be judged against a highly determinate target.

*Table 6. Question-design rubric as risk control.*

| Criterion | Risk controlled |
|---|---|
| *Source fidelity* | Prevents vague attribution; answers are validated against a named source work, passage, or stated scholarly scope. |
| *Determinacy* | Prevents unbounded questions; items are constrained to one correct reading under the specified school or accepted-answer set. |
| *Scholarly demand alignment* | Protects tier interpretability; each item is verified to match both its source stratum and its expected cognitive demand. |
| *Terminological precision* | Prevents semantic ambiguity; review verifies discipline-appropriate vocabulary over general Arabic synonyms. |
| *Linguistic clarity* | Reduces wording noise; Arabic phrasing is tightened to remove hidden context and unclear scope. |
| *Distractor quality* | Prevents cue-based MCQ success; distractors must be plausible under the topic but definitively incorrect under the keyed position. |
| *Pluralism control* | Stabilises scoring amid valid disagreement; items explicitly specify the relevant source, author, or school. |

## 4.3 Refinement audit and gold-standard correction

Bounded question design is necessary but insufficient if the reference answers or item constraints contain residual defects. To stabilise the gold standard prior to any formal model evaluation, the dataset underwent a rigorous refinement audit. This pre-evaluation validation phase ensured that downstream benchmarking scores would reflect genuine model capabilities rather than dataset noise.

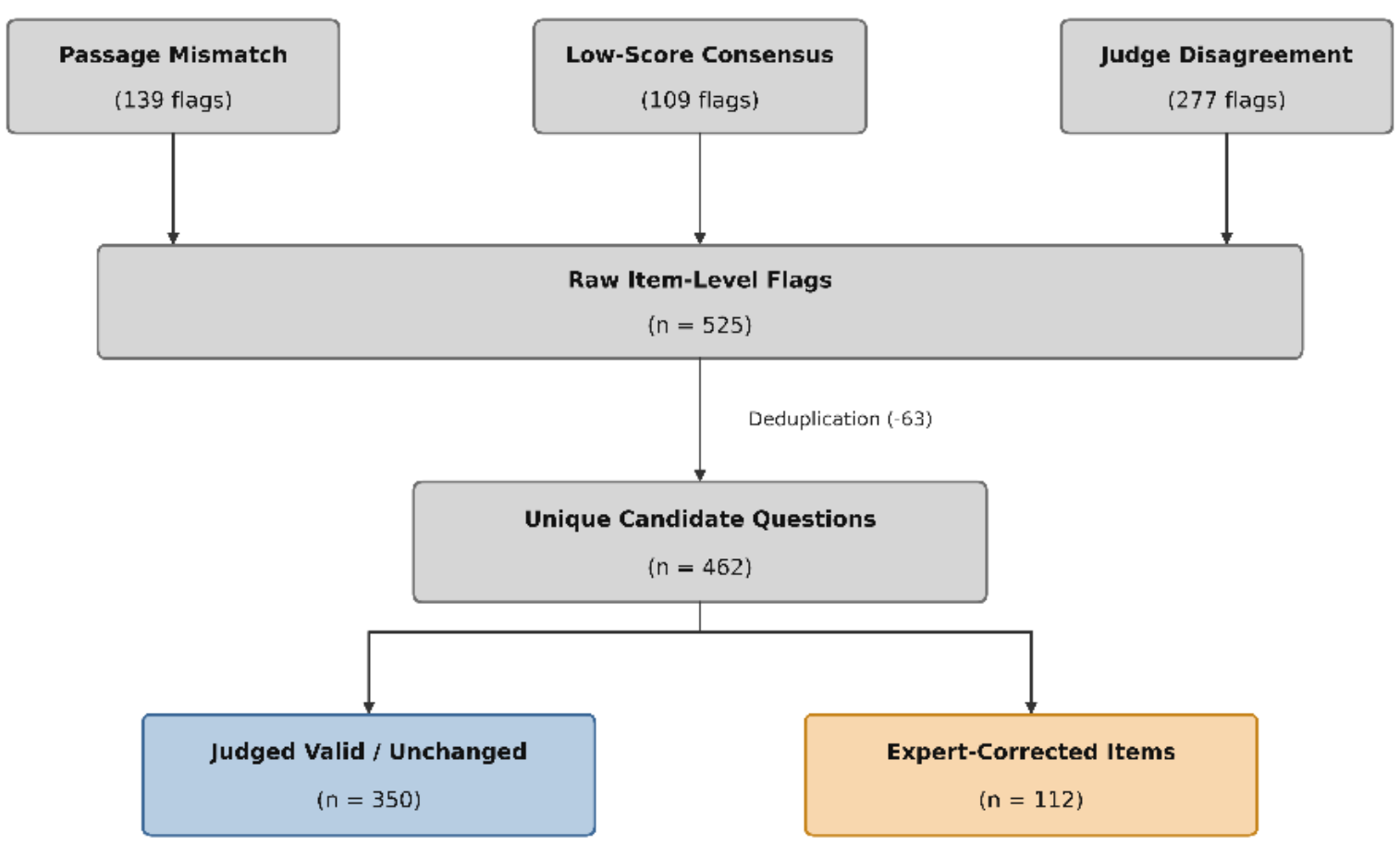

*Figure 3. Refinement audit triage pipeline. Automated diagnostic signals isolated 462 unique candidate questions. Expert re-review confirmed 112 items requiring correction, while validating 350 items as structurally sound without modification.*

**Triage and screening protocol.** The audit utilised diagnostic outputs from an initial, zero-shot pre-release benchmarking run strictly as screening signals to route candidate questions to expert re-review. In this diagnostic run, MCQ responses were evaluated via a deterministic exact-match script, while COMP and KNOW responses were scored using a reference-guided LLM-as-a-judge protocol utilising two judges (GPT-5.2 and Gemini-2.5-Flash). This methodology directly mirrors the formal evaluation protocols detailed later in Usage Illustrations, ensuring that the triage signals accurately reflect the benchmark's intended downstream environment.

As illustrated in Figure 3, three automated heuristics were applied to these diagnostic outputs: passage mismatches in COMP items, low-score consensus across frontier models (GPT-5.1 and Gemini-3-Pro), and high disagreement (absolute difference ≥ 0.5) between the two LLM judges. These signals generated 525 raw flags, which were deduplicated to isolate a targeted pool of 462 unique candidate questions.

**Correction outcomes.** All 462 candidates underwent manual re-assessment by the expert panel. The screening signals were treated exclusively as triage evidence, not as definitive proof of a defect. Upon review, 350 items were judged valid and left unchanged. The experts confirmed and corrected 112 unique questions (3.2% of the released corpus), comprising 58 KNOW, 29 COMP, and 25 MCQ items (Table 7).

*Table 7. Refinement audit outcomes and reliability metrics.*

| *Metric* | Value |
|---|---|
| *Unique expert-*corrected questions | 112 |
| *Share of released corpus corrected* | 3.2% |
| *KNOW / COMP / MCQ corrections* | 58 / 29 / 25 |
| *Deleted items* | 0 |
| *Final corpus size* | 3,465 |
| *Interval Krippendorff's α* | 0.81 |

The corrections successfully resolved ambiguities across question stems, MCQ keys, distractors, and open-ended scoring boundaries (Table 8). Inter-rater agreement during this refinement process yielded an interval Krippendorff's α of 0.81, indicating strong consensus on the required partial-quality adjustments.

*Table 8 Correction-fix-type frequencies applied during the audit.*

| Fix type | Count | What was corrected |
|---|---|---|
| *Question rewording* | 82 | Ambiguous wording, unclear scope, or hidden-context dependence. |
| *Gold-answer revision* | 78 | Reference answer clarified, completed, or corrected. |
| *KNOW scoring-rule adjustment* | 58 | no_requested_items threshold or accepted-answer set revised. |
| *MCQ distractor revision* | 13 | Distractor plausibility, ambiguity, or cueing corrected. |
| *MCQ key correction* | 12 | Keyed correct option reassigned. |
| *COMP passage-excerpt revision* | 3 | Passage boundary was adjusted to support the question/answer sufficiently. |

# 5 Usage Illustration

The following usage illustration demonstrates how IslamicTurathBench can be used for reproducible baseline evaluation of large language models over the classical Islamic scholarly tradition. The experiment is not intended to establish a permanent model ranking. Rather, it shows how the dataset supports comparison across systems, task formats, scholarly-demand tiers, and disciplines under a controlled prompting and scoring protocol.

## 5.1 Subject systems

We evaluate ten systems spanning four groups: frontier proprietary generalists, open-weight DeepSeek models, Arabic-specialised systems, and Islamic-domain deployed systems (summarised in Table 9). ALLaM-2-34B was evaluated through a blind operator-run protocol with gold labels and reference answers withheld. Fanar-Sadiq and Islamic-RAG are evaluated as deployed systems, so their scores measure end-to-end behaviour rather than parametric recall alone.

*Table 9. Subject systems evaluated in IslamicTurathBench. The table lists the ten evaluated systems, grouped by system type, with parameter information and deployment notes for the evaluation window from 11 December 2025 to 3 March 2026. n/d = not disclosed; n/a = deployed system.*

| Group | Model/System | Params |
|---|---|---|
| **Frontier** | GPT-5.1 (OpenAI, 2025)[25] | n/d |
| **Frontier** | Gemini-3-Pro (Google DeepMind, 2025)[26] | n/d |
| **Open-weight** | DeepSeek-V3 (DeepSeek-AI, 2024)[27] | 671B† / 37B† |
| **Open-weight** | DeepSeek-R1 (DeepSeek-AI, 2025)[28] | 671B† / 37B† |
| **Arabic** | ALLaM-2-34B* (Bari et al., 2024)[29] | 34B |
| **Arabic** | ALLaM-2-7B (SDAIA — National Center for Artificial Intelligence, 2025)[30] | 7B |
| **Arabic** | Jais-2-70B-Chat (Anwar et al., 2025)[31] | 70B |
| **Arabic** | Fanar-C-2-27B (FANAR TEAM et al., 2026)[32] | 27B‡ |
| **Islamic** | Fanar-Sadiq (Abbas et al., 2026)[33] | n/a |
| **Islamic** | Islamic-RAG | n/a |

* ALLaM-2-34B was evaluated through a blind operator-run protocol with gold labels and reference answers withheld. Gemini-3-Pro refers to the preview deployment evaluated during the stated window. † MoE: 671B total, 37B active/token. ‡ Fanar-C-2-27B API label; public model: Fanar-2-27B-Instruct.

## 5.2 Experimental Setup

All systems are evaluated under a zero-shot protocol. Prompts are task-specific: MCQ questions require a single Arabic option letter; COMP and KNOW questions require an open-ended Arabic answer and a verbalised confidence score from 0 to 100%. All runs use temperature 0, a maximum of 2,000 tokens, and up to three retries for failed calls. Complete rendered prediction prompts are provided in the Supplementary Materials (Section S5).

## 5.3 Scoring pipeline

MCQ questions are scored by exact match against the gold option label. COMP and KNOW questions are scored with a reference-guided LLM-as-a-judge pipeline [34,35]. The judge receives the question, the gold reference answer, the model's answer, and, for COMP, the source passage. It returns a JSON object containing a scalar score in [0, 1] at 0.1 increments and an explanation justifying the score. The open-ended scoring prompts are provided in the Supplementary Materials (Section S6).

We screened five candidate judge models on 20 held-out KNOW questions under Islamic-studies expert review. GPT-5.2 and Gemini-2.5-Flash were retained as production judges. The model gemini-3.1-pro-preview was reserved as head judge for high-disagreement adjudication. For each COMP or KNOW answer, the two production judges score independently. If the absolute difference between their scores satisfies $|j1 - j2| \leq 0.2$, the final score is their mean. If $|j1 - j2| > 0.2$, the response is escalated to the head judge for a final score. This occurred for 18.4% of open-ended responses.

Production-judge agreement was measured with Krippendorff's α for interval data, using squared-distance disagreement on the 0–1 score scale. Agreement was high for both COMP ($\alpha = 0.937$; $n = 3,966$) and KNOW ($\alpha = 0.936$; $n = 7,662$), with a pooled overall agreement $\alpha = 0.944$.

## 5.4 Evaluation metrics

The overall ISTB score is a cell-balanced macro-average. For each model, we first compute the mean score within each populated design cell. A design cell is defined by one discipline, one task format, and one scholarly-demand tier. We then compute the mean of these cell scores to obtain the overall ISTB score. Hence, let Sd,t,k(m) denote the mean score for model m in discipline d, task format t, and scholarly demand tier k. ISTB is the equal-cell macro-average over all populated Scholarly demand × Discipline × Format cells:

$$ISTB(m) = \frac{1}{|C|} \sum_{(d,t,k)\in C} S_{d,t,k}(m)$$

where C denotes the set of populated discipline × task-format × scholarly-demand cells, and S_{d,t,k}(m) denotes the mean score of model m in discipline d, task format t, and scholarly-demand tier k.

In other words, ISTB first averages scores within each cell, then averages across cells. This gives each populated discipline–task-format–scholarly-demand cell equal weight, regardless of how many individual questions it contains. The design prevents larger item groups, especially MCQ, from dominating the headline score simply because they contain more questions.

We also report marginal scores along each benchmark axis. A task-format marginal averages all cells with the same task format across disciplines and scholarly-demand tiers. A scholarly-demand marginal averages all cells from the same tier across disciplines and task formats. A discipline marginal averages all cells from the same discipline across task formats and scholarly-demand tiers.

**Calibration metrics.** For COMP and KNOW, we evaluate calibration by comparing the model's verbalised confidence to the final continuous judge score in [0, 1] [36]. Calibration is therefore treated as confidence–quality alignment over graded answers, not as Bernoulli calibration against binary correctness. MCQ calibration is not reported because MCQ prompts do not elicit verbalised confidence to facilitate exact-match evaluation. The primary calibration metric is RMSCE, root mean squared calibration error [37]. We also report Spearman's ρ, which measures the rank association between model confidence and judged answer quality. Lower RMSCE indicates closer alignment between confidence and quality; higher Spearman's ρ indicates that confidence better tracks relative answer quality.

## 5.5 Baseline performance across benchmark axes

Table 10 reports the baseline leaderboard with task-format marginals. Gemini-3-Pro obtains the highest aggregate ISTB score (0.889), followed by GPT-5.1 (0.826). The middle of the ranking is more compressed: DeepSeek-R1, DeepSeek-V3, ALLaM-2-34B, Fanar-Sadiq, Islamic-RAG, and Jais-2-70B-Chat all fall within a relatively narrow aggregate band.

*Table 10. ISTB Baseline leaderboard with task-format marginals. Zero-shot system performance on IslamicTurathBench. ISTB is the aggregate benchmark score; MCQ, COMP, and KNOW report task-format marginals.*

| Rank | System/model | ISTB | 95% CI | MCQ | COMP | KNOW |
|---|---|---|---|---|---|---|
| 1 | Gemini-3-Pro | 0.8886 | 0.8627–0.9144 | 0.9458 | 0.9234 | 0.7964 |
| 2 | GPT-5.1 | 0.8261 | 0.7911–0.8612 | 0.8521 | 0.9421 | 0.6842 |
| 3 | DeepSeek-R1 | 0.7763 | 0.7320–0.8205 | 0.8095 | 0.9260 | 0.5932 |
| 4 | DeepSeek-V3 | 0.7651 | 0.7213–0.8088 | 0.7982 | 0.9145 | 0.5825 |
| 5 | ALLaM-2-34B | 0.7584 | 0.7142–0.8027 | 0.7831 | 0.8709 | 0.6214 |
| 6 | Fanar-Sadiq | 0.7456 | 0.7031–0.7881 | 0.8308 | 0.7991 | 0.6171 |
| 7 | Islamic-RAG | 0.7449 | 0.7042–0.7857 | 0.8297 | 0.8102 | 0.5949 |
| 8 | Jais-2-70B-Chat | 0.7296 | 0.6864–0.7729 | 0.7999 | 0.8512 | 0.5377 |
| 9 | Fanar-C-2-27B | 0.6337 | 0.5778–0.6897 | 0.7168 | 0.8111 | 0.3734 |
| 10 | ALLaM-2-7B | 0.5433 | 0.4915–0.5951 | 0.7094 | 0.5822 | 0.3383 |

**Task format.** Averaged across systems, COMP had the highest mean marginal score (0.8431), followed by MCQ (0.8075) and KNOW (0.5739) (Fig. 4). These differences show why IslamicTurathBench reports task-format marginals rather than only a single aggregate score: answer conditions change the measured performance profile. COMP is the strongest format for six of the ten systems, including GPT-5.1, both DeepSeek systems, ALLaM-2-34B, Jais-2-70B-Chat, and Fanar-C-2-27B. This pattern suggests that, for many systems, passage grounding provides stronger support than answer options alone.

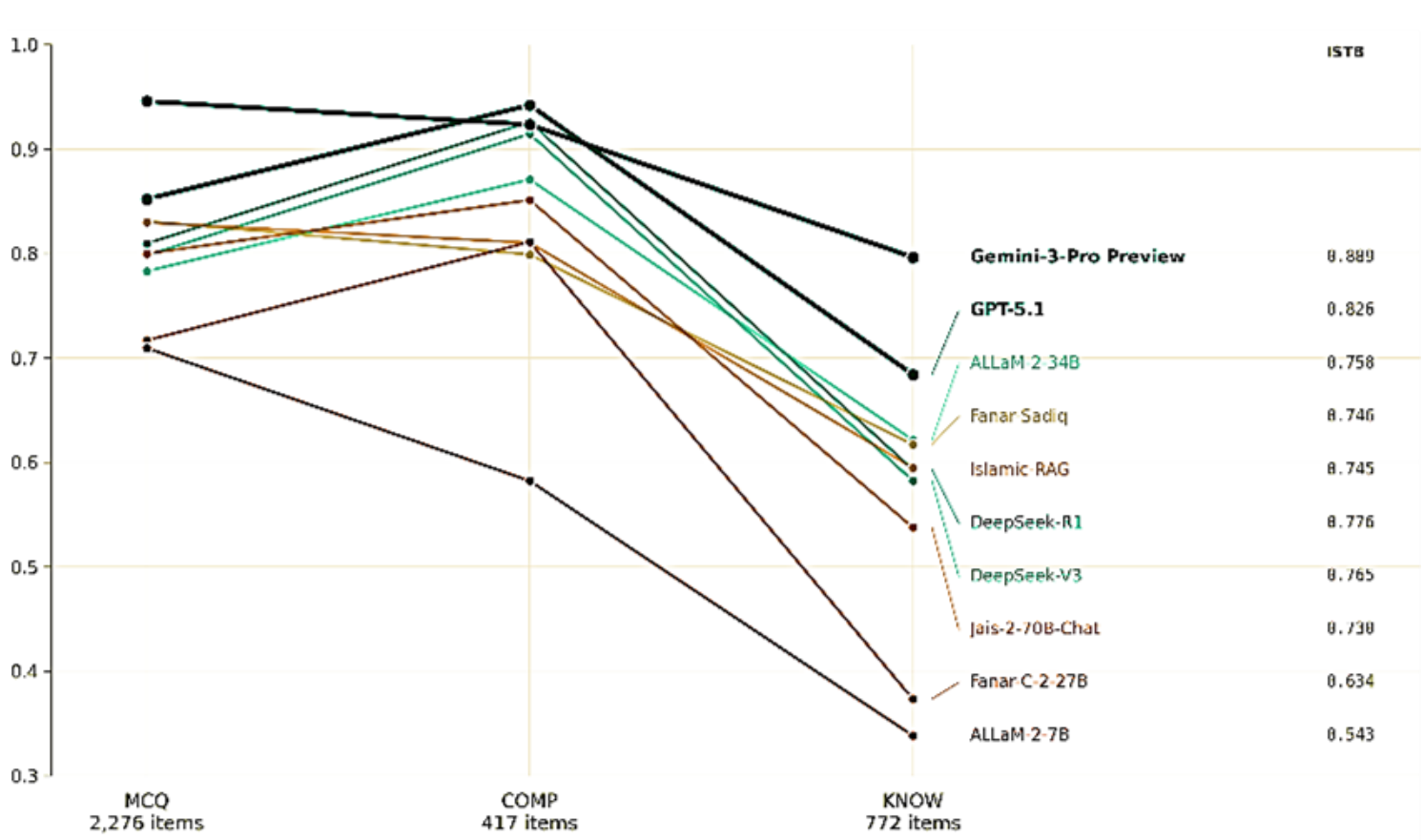

*Figure 4. Task-format effects on system performance. Mean performance by task format for each evaluated system in IslamicTurathBench. Each line represents one system across MCQ, COMP, and KNOW; colours encode overall ISTB rank, and right-side labels report each system's aggregate ISTB score. The figure illustrates that system performance is task-format dependent, with most systems scoring higher under answer-option or passage-supported conditions than under closed-book open-ended answering.*

**Scholarly demand.** Performance also varied by scholarly demand level. Averaged across systems, Beginner items had the highest mean score (0.8193), followed by Intermediate items (0.7322) and Advanced items (0.6716). This pattern is consistent with the benchmark design, in which scholarly demand tiers increase the required scholarly operation from direct recall and local inference toward comparison, evidential reasoning, application, and synthesis. Task-format differences persist across scholarly-demand levels. Table 11 reports the average task format × scholarly demand scores across the ten evaluated systems. KNOW remains below the supported task formats at all three levels, while COMP is strongest at the Beginner and Intermediate levels and declines more sharply at Advanced. This illustrates how the benchmark can separate scholarly-demand effects from answer-condition effects.

*Table 11. Average task-format × scholarly demand scores across the ten evaluated systems.*

| Task format | Beginner | Intermediate | Advanced |
|---|---|---|---|
| **MCQ** | 0.847 | 0.803 | 0.772 |
| **COMP** | 0.932 | 0.862 | 0.734 |
| **KNOW** | 0.679 | 0.534 | 0.509 |

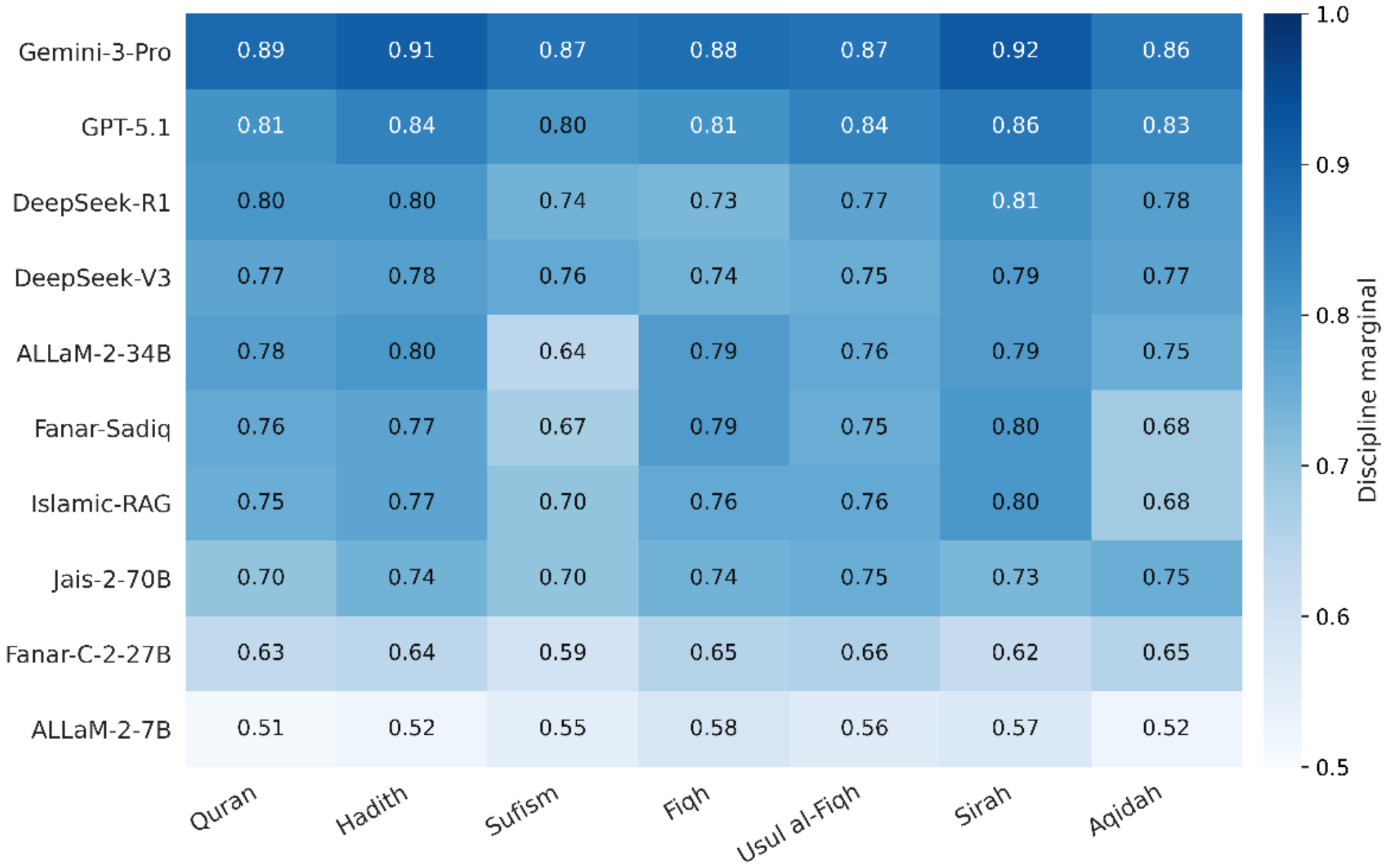

*Figure 5. Discipline-level closed-book performance gap by system. Heatmap showing, for each system and discipline, the difference between closed-book open-ended performance (KNOW) and the stronger of the two supported task formats, MCQ or COMP. Values closer to zero indicate smaller task-format dependence, whereas more negative values indicate a larger drop when systems answer without answer options or a supplied passage. The gap is descriptive and should be interpreted alongside the corresponding discipline-level performance marginals.*

**Discipline**. Discipline-level marginals provide a further diagnostic view of benchmark performance (Fig. 5). Averaged across systems, discipline means ranged from 0.7677 for Prophetic biography and 0.7571 for Hadith sciences to 0.7276 for Islamic theology and 0.7012 for Sufism. These values should be interpreted descriptively rather than causally. Discipline-level differences may reflect source genre, technical vocabulary, question composition, task-format mix, and the degree to which the underlying material is represented in model training or retrieval environments.

*Table 12. Confidence–quality alignment for COMP and KNOW.*

| System/model | COMP RMSCE | COMP Spearman's ρ | KNOW RMSCE | KNOW Spearman's ρ |
|---|---|---|---|---|
| **Gemini-3-Pro** | 0.065 | 0.108 | 0.185 | 0.137 |
| **GPT-5.1** | 0.044 | 0.497 | 0.267 | 0.502 |
| **DeepSeek-R1** | 0.030 | 0.577 | 0.344 | 0.255 |
| **DeepSeek-V3** | 0.061 | 0.487 | 0.372 | 0.260 |
| **ALLaM-2-34B** | 0.148 | 0.402 | 0.373 | 0.197 |
| **Fanar-Sadiq** | 0.165 | 0.588 | 0.369 | 0.267 |
| **Islamic-RAG** | 0.157 | 0.519 | 0.385 | 0.231 |
| **Jais-2-70B-Chat** | 0.082 | 0.090 | 0.424 | 0.242 |
| **Fanar-C-2-27B** | 0.169 | 0.260 | 0.587 | 0.059 |
| **ALLaM-2-7B** | 0.381 | 0.198 | 0.630 | 0.056 |

## 5.6 Confidence–quality alignment

Table 12 reports RMSCE and Spearman's ρ for COMP and KNOW. Across systems, mean confidence is consistently higher than mean judged quality. GPT-5.1 has the strongest confidence ranking on KNOW responses, with Spearman's ρ = 0.502 between confidence and judged score, but its mean confidence still exceeds its mean judged quality by 0.247. The largest confidence–quality gaps occur among the lower-scoring KNOW systems. These results illustrate that confidence values should not be interpreted as direct substitutes for judged answer quality, especially in closed-book open-ended settings.

### 5.7 Matched-subset scholarly reference comparison

*Table 13. Matched-subset human reference comparison. Micro-averaged scores over answered questions on the matched human-reference subset. The selected systems illustrate the task-format dependence of human–model comparisons. The human reference panel is a contextual generalist scholarly baseline rather than a specialist performance ceiling.*

| System/model | ISTB | MCQ | COMP | KNOW |
|---|---|---|---|---|
| **Human reference panel** | **0.754** | **0.884** | **0.632** | **0.748** |
| **Gemini-3-Pro** | 0.900 | 1.000 | 0.928 | 0.771 |
| **GPT-5.1** | 0.818 | 0.870 | 0.917 | 0.667 |
| **ALLaM-2-34B** | 0.787 | 0.830 | 0.827 | 0.704 |
| **Fanar-Sadiq** | 0.768 | 0.891 | 0.747 | 0.666 |
| **Islamic-RAG** | 0.772 | 0.891 | 0.762 | 0.664 |
| **DeepSeek-R1** | 0.767 | 0.783 | 0.909 | 0.611 |

The matched human-reference subset provides a contextual comparison point for interpreting model scores on the same questions answered by Islamic Studies experts. Because the subset contains a limited number of questions per design cell, scores are reported using micro-averaging over answered questions rather than full macro-cell aggregation. The human reference panel should be interpreted as a generalist scholarly baseline, not as a specialist ceiling or as an evaluation of individual scholars.

Table 13 reports selected matched subset results. Several systems exceed the human reference panel in aggregate score, but the comparison is strongly task-format dependent. Model advantages are most visible in COMP, where responses are supported by a supplied passage. KNOW gives a different view: among the selected systems shown, only Gemini-3-Pro exceeds the human panel on closed-book open-ended questions. The human panel also has a different task profile from most models, with comparatively stronger performance on KNOW than on COMP.

This usage illustration shows why aggregate human–model comparisons should be interpreted with task-format control. In IslamicTurathBench, the matched subset is therefore useful not as a single ranking device, but as a reference layer for comparing answer-option recognition, passage-supported answering, and closed-book open-ended response generation.

All baseline evaluations reported in this manuscript were completed before the public release of IslamicTurathBench v1.0. Results obtained after public release should be interpreted in light of possible benchmark exposure through model training, retrieval systems, or other data-access mechanisms. Future evaluations should report the benchmark version, model version, and evaluation date.

## 6 Data Availability

IslamicTurathBench (ISTB) v1.0 is publicly available on Zenodo under DOI 10.5281/zenodo.20674930[24]. The deposit contains three UTF-8 task collections (istb_mcq, istb_comp, and istb_know) in JSON and CSV format, comprising 3,465 question–answer items in total, together with documentation, source-work metadata, an integrity manifest with SHA-256 checksums and per-task schemas, a reference loader, build-level metadata, aggregate corpus statistics, and the aggregated scholarly human-reference layer. The dataset is released under a Creative Commons Attribution 4.0 International (CC BY 4.0) licence.

# 7 Code Availability

The evaluation code and aggregate result files supporting the analyses, tables, and figures reported in this manuscript are publicly available at https://github.com/GabenS99/IslamicTurathBench_Evaluation. The benchmark dataset is archived separately on Zenodo and is not duplicated in the code repository. Raw model responses and intermediate LLM-judge output files are not included in the public release.

## Author Contributions

Shahd Gaben: Writing, Methodology, Data Curation, Software, Formal Analysis, Validation, Visualization, Project Administration. Heba Sbahi: Data Curation, Investigation, Validation, Methodology, Review & Editing. Samer Rashwani: Conceptualization, Methodology, Investigation, Resources, Validation, Review & Editing. Abdessalam Bouchekif: Conceptualization, Methodology, Validation, Review & Editing. Mutaz Al-Khatib: Conceptualization, Methodology, Investigation, Resources, Validation, Review & Editing. Emad Mohamed: Supervision, Methodology, Review & Editing. Somaya Eltanbouly: Methodology, Review & Editing. Mohammed Ghaly: Conceptualization, Methodology, Supervision, Resources, Validation, Review & Editing, Funding.

## Competing Interests

The authors declare no competing interests.

## Funding

This study was conducted as part of a project funded by the ARG grant (ARG01-0524-230318), awarded by the Qatar Research, Development, and Innovation Council (QRDI).

## Ethics statement

The human reference component involved anonymous aggregate responses from Islamic Studies experts for the purpose of constructing a scholarly reference baseline. Participants were informed about the purpose of the task, the expected use of their responses in aggregate form, and the exclusion of personally identifying information from the public dataset. Raw individual responses and evaluator identities are not included in the released dataset.

# Supplementary Materials

.

***Prompt and translation fidelity. T****he Arabic prompts are reproduced from the dataset-construction and evaluation workflows verbatim. The English translations are provided for documentation and were not supplied to the systems during generation or evaluation. English translations of benchmark examples are provided only for reader accessibility and are not part of the released Arabic benchmark.*

## 1. Dataset-related materials

### S1. Scholarly-demand tier examples

*Table S1. Scholarly-demand examples from Jurisprudence. The examples show how the tier label changes the required operation, not merely the source title.*

| Tier | Example question |
|---|---|
| Beginner | متى يجب الفطر على الصائم؟<br>When is a fasting person obligated to break their fast? |
| Intermediate | هل يجوز توكيل المرأة بخلع زوجة زوجها أو طلاقها، وفق الأصح في المذهب الشافعي؟<br>According to the stronger view in the Shāfiʿī school, may a woman be appointed as agent to obtain or pronounce the khulʿ or divorce of her husband's other wife (her co-wife)? |
| Advanced | لو ولدت امرأة ولدين ملزقين لهما رأس واحد وأربع أرجل وأربع أيد وفرجان، فما العلة في اعتبار الولدين الملزقين بمنزلة الاثنين المستقلين في المواريث؟<br>If a woman gives birth to conjoined twins with one head, four legs, four hands, and two distinct genital organs, what is the juristic rationale (ʿilla) for treating them as two independent heirs in matters of inheritance? |

### S2. Task-format example

Table S2 gives a parallel Advanced example in Principles of Jurisprudence. The questions share the same discipline and Advanced scholarly-demand tier but vary by task format.

*Table S2. Advanced Principles of Jurisprudence examples illustrating the task-format gradient. MCQ provides answer options, COMP provides a source passage, and KNOW provides no external support.*

| Format | Example |
|---|---|
| MCQ | إذا ورد عن صحابيٍّ قولٌ أو فعلٌ في أمرٍ تعبديٍّ لا مدخل للقياس فيه، فما هو الموقف منه وفقاً لما استنبطه محققو الشافعية من قول الإمام الشافعي الجديد؟<br>أ. .يُعدُّ حجةً يجب المصير إليها، لظهور كونه توقيفيًا، وهذا الموضع مستثنىً من عموم القول الجديد<br>ب. .يُردُّ عملاً بإطلاق القول الجديد القاضي بأن مذهب الصحابي ليس بحجة، ولا يُصار إلى الاستثناء<br>ج. .يُعتبر دليلاً مرجوحًا، لا يُعمل به إلا عند انعدام الأدلة من الكتاب والسنة والإجماع والقياس الجلي<br>د. .يُحمل على أنه من مسائل القول القديم المهجور، فلا يُلتفت إليه في مقام الإفتاء والعمل<br><br>English translation:<br>If a statement or act is transmitted from a Companion regarding a purely devotional matter in which analogical reasoning (qiyās) has no place, what is the position on it according to what the leading Shāfiʿī scholars derived from Imām al-Shāfiʿī's later doctrine (al-qawl al-jadīd)?<br>A. It counts as binding authority that must be followed, since its devotional character makes it manifestly grounded in revelation (tawqīfī); this case is an exception to the general rule of the later doctrine.<br>B. It is rejected, in accordance with the unrestricted form of the later doctrine, which holds that the position of a Companion is not authoritative; no exception is admitted.<br>C. It is treated as a weaker form of evidence, acted upon only when no proof can be found in the Qurʾān, the Sunna, scholarly consensus (ijmāʿ), or manifest analogical reasoning (al-qiyās al-jalī). |

| Format | Example |
|---|---|
| | D. It is taken to belong to the abandoned earlier doctrine (al-qawl al-qadīm) and is therefore disregarded in legal opinion (iftāʾ) and practice. |
| COMP | النص) :مسألة :الأكثر أن حصول الشرط الشرعي ليس شرطا في صحة التكليف .(ش :أي بالمشروط، بل يصح التكليف بالمشروط حالة عدم الشرط، خلافا لأهل الرأي والمراد بـ )الشرط الشرعي (ما يتوقف عليه صحة الشيء شرعا كالوضوء للصلاة فخرج ما يتوقف عليه وجوده عقلا كالتمكن من الأداء الزائل بالنوم والفهم من الخطاب الزائل بالغفلة والنسيان فإن حصوله شرط في صحة التكليف، وقد سبق، وقد استشكل الفرق بينهما وبين المسألة السابقة في مقدمة الواجب، فإنها إذا وجبت وجب تحصيل الشرط فما فائدة ذكر هذه المسألة؟ قلت :الكلام في حصول الشرط الشرعي بالنسبة إلى الصحة، فعندنا لا يتوقف صحة التكليف على حصوله، ومسألة المقدمة بالنسبة إلى الواجب نفسه إذا توقف على أمر آخر من شرط أو غيره، هل يوجب المقدم، فهما غيران لا تعلق لأحدهما بالآخر.<br><br>السؤال :حرِّر من خلال النص معيارَ التفريق بين ما يتوقف عليه الشيء شرعًا وما يتوقف عليه عقلًا، ثم طبِّق ذلك على حالتي النوم والغفلة من جهة صحة التكليف.<br><br>English translation:<br>Passage: (Issue: the majority view is that the obtaining of a legal precondition is not itself a condition for the validity of legal obligation [taklīf].) Commentary: that is, with respect to the conditioned act; rather, legal obligation regarding a conditioned act remains valid even when the precondition is not met, contrary to the proponents of independent juristic reasoning (ahl al-raʾy). By "legal precondition" is meant that on which the legal validity of an act depends, such as ablution (wuḍūʾ) for prayer. This excludes what something depends on rationally for its very occurrence, such as the capacity to perform an act (which lapses during sleep) and the ability to comprehend an address (which lapses through inattention or forgetfulness); the obtaining of such conditions is required for the validity of legal obligation, as established earlier. It has been objected that the distinction between this issue and the earlier issue of "the prerequisite of the obligatory act" (muqaddimat al-wājib) is unclear: if the obligatory act is mandated, then the precondition is mandated as well, so what is the benefit of stating this separate issue? I reply: the discussion here concerns the obtaining of the legal precondition as it relates to validity, and our view is that the validity of legal obligation does not depend on it; the issue of the prerequisite, by contrast, concerns the obligatory act itself when it depends on some other matter (a precondition or otherwise), and whether the prerequisite is itself thereby mandated. The two are distinct issues with no bearing on one another.<br><br>Question: Drawing on the passage, formulate the criterion that distinguishes what a thing depends on legally from what it depends on rationally, then apply this criterion to the cases of sleep and inattention with respect to the validity of legal obligation. |
| KNOW | بناء على ما ذكره الزركشي في "تشنيف المسامع" ما رأي الأكثرين من الشافعية في حكم المجاز بالنسبة لإثبات الأحكام؟ وضح سبب رأيهم.<br><br>English translation:<br>Based on what al-Zarkashī states in Tashnīf al-Masāmiʿ, what is the position held by the majority of the Shāfiʿī scholars on the use of figurative language (majāz) as a basis for establishing legal rulings? Explain the reasoning behind their view. |

## S3. Question-authoring and validation rubric

The internal construction rubric separated question drafting from validation. AI-assisted prompting produced candidate questions. Expert review determined whether a candidate became a benchmark question. Table S3 summarizes the validation criteria used during expert review.

*Table S3. Question-validation criteria used during expert review.*

| Criterion | Operational test |
|---|---|
| Source fidelity | The answer is traceable to the assigned source work or stated scholarly position. |
| Determinacy | The question has one correct answer under the specified source, scholar, or school constraint. |
| Scholarly-demand alignment | The required operation matches the assigned Beginner, Intermediate, or Advanced tier. |
| Terminological precision | The wording uses discipline-appropriate Islamic Studies terminology. |
| Linguistic clarity | The Arabic question is clear and does not depend on hidden context. |

| Criterion | Operational test |
|---|---|
| Distractor quality | For MCQ, distractors are plausible but definitively incorrect under the stated source or scholarly position. |
| Pluralism control | Legitimate disagreement is constrained by source, school, or named scholar. |

## S4. Representative question-drafting prompts

Gemini 2.5 Pro was used to generate candidate questions from the assigned source texts. The prompts specified the target tier and requested direct, answerable Arabic questions. Candidate drafts were then rewritten during human review. Reviewers edited wording to match the scholarly register of the source material, corrected terminology, removed ambiguous phrasing, and constrained questions where disagreement could produce multiple valid answers. The paper therefore treats AI assistance as a drafting aid, not as an annotation authority.

Representative Arabic drafting prompts. The following prompts are representative examples used to generate initial MCQ question drafts at the Beginner, Intermediate, and Advanced scholarly-demand tiers.

### Beginner

**Arabic prompt**

أنت متخصص في العلوم الشرعية، ومطلوب منك وضع أسئلة معيارية على شكل مسائل عملية تفيد الناس من نص الكتاب المقدم إليك. يجب أن تركز الأسئلة على الحصول على معلومات مباشرة (أسماء، تعاريف، أماكن بعينها). رتّب الأسئلة والأجوبة في ملف إكسل. اذكر الخيارات، وضع خانة للإجابة الصحيحة، واذكر رمز الجواب الصحيح. قم بصياغة أسئلة مباشرة ومفهومة دون الرجوع إلى النص المعطى، ودون الإشارة إلى النص المعطى (لا تقل: كما ورد في النص، أو: حسب النص). نوّع صياغة الأسئلة واجعلها على شكل أحجية علمية. اجعل الأسئلة مفهومة دون الرجوع إلى النص.

**English translation**

You are a specialist in the Islamic sciences, and you are asked to draft benchmark questions as practical cases that are useful to readers from the source text provided to you. The questions should focus on direct information (names, definitions, and specific places). Arrange the questions and answers in an Excel sheet. Provide the options, add a field for the correct answer, and record the symbol of the correct option. Write direct, clear questions that can be understood without returning to the passage, and do not refer to the passage explicitly (do not say "as stated in the text" or "according to the text"). Vary the phrasing and present the questions as scholarly riddles. The questions should remain understandable without referring back to the passage.

### Intermediate

**Arabic prompt**

أنت متخصص في العلوم الشرعية، ومطلوب منك وضع أسئلة معيارية على شكل مسائل عملية تفيد الناس من نص الكتاب المقدم إليك. يجب أن تركز الأسئلة على اختبار القدرة على التفريق، والمقارنة، وذكر الأدلة. رتّب الأسئلة والأجوبة في ملف إكسل. اذكر الخيارات، وضع خانة للإجابة الصحيحة، واذكر رمز الجواب الصحيح. قم بصياغة أسئلة مباشرة ومفهومة دون الرجوع إلى النص المعطى. نوّع صياغة الأسئلة واجعلها على شكل أحجية أو فزورة علمية.

**English translation**

You are a specialist in the Islamic sciences, and you are asked to draft benchmark questions as practical cases that are useful to readers from the source text provided to you. The questions should focus on testing the ability to distinguish, compare, and cite evidence. Arrange the questions and answers in an Excel sheet. Provide the options, add a field for the correct answer, and record the symbol of the correct option. Write direct, clear questions that can be understood without returning to the supplied passage. Vary the phrasing and present the questions as scholarly riddles or puzzles.

### Advanced

#### Arabic prompt

أنت متخصص في العلوم الشرعية، ومطلوب منك وضع أسئلة معيارية على شكل مسائل عملية تفيد الناس من نص الكتاب المقدم إليك. يجب أن تركز الأسئلة على اختبار القدرة على التفريع والتفصيل، والتعليل والتحليل، والتطبيق والانتقال من النظر إلى التطبيق على حالات بعينها. رتّب الأسئلة والأجوبة في ملف إكسل. اذكر الخيارات، وضع خانة للإجابة الصحيحة، واذكر رمز الجواب الصحيح. قم بصياغة أسئلة مباشرة ومفهومة دون الرجوع إلى النص المعطى.

#### English translation

You are a specialist in the Islamic sciences, and you are asked to draft benchmark questions as practical cases that are useful to readers from the source text provided to you. The questions should focus on testing the ability to derive subsidiary rulings, elaborate details, supply legal justification and analysis, and move from theoretical reflection to application in concrete cases. Arrange the questions and answers in an Excel sheet. Provide the options, add a field for the correct answer, and record the symbol of the correct option. Write direct, clear questions that can be understood without returning to the supplied passage.

## 2. Evaluation-related materials

### S5. Default prediction prompt templates

All subject-system prompts follow the zero-shot protocol described in the main article. This section reproduces one complete rendered example per task format. The discipline-specific role instruction is instantiated for this example; in the actual run, the discipline token is substituted per question. The same protocol is used across all scholarly-demand tiers.

*Table S4. Default Arabic prediction-prompt structure. Full rendered examples are reproduced below.*

| Step | Component | Content | Task scope |
|---|---|---|---|
| 1 | Role instruction | Discipline-specific Arabic expert-role instruction | All tasks |
| 2 | Task instruction | Task-type directive for MCQ, COMP, or KNOW | All tasks |
| 3 | Question and inputs | MCQ: question + four options; COMP: passage + question; KNOW: question only | Task-specific |
| 4 | Required output format | MCQ: one option letter; COMP/KNOW: Arabic answer + confidence value | Task-specific |

### MCQ

#### Arabic rendered prompt

أنت خبير متخصص في أصول الفقه، تتمتع بمعرفة عميقة ودقيقة في هذا المجال.

مهمتك هي اختيار الإجابة الصحيحة من بين الخيارات المتاحة.

السؤال: نصوص القرآن من جهة ورودها وثبوتها ونقلها عن الرسول، تعتبر:

الخيارات:
أ) ظنية
ب) بعضها قطعي وبعضها ظني.
ج) كل الأجوبة خطأ
د) قطعية

تعليمات الإجابة:
- أجب بحرف عربي واحد فقط من الخيارات التالية: أ، ب، ج، د
- لا تكتب أي شرح أو تفسير إضافي

- لا تكتب كلمات إضافية غير الحرف

**English translation**

You are a specialist in Principles of Jurisprudence with deep and precise expertise in this field.

Your task is to select the correct answer from the available options.

Question: The texts of the Qurʾān, viewed from the perspective of their transmission, attestation, and narration from the Prophet, are considered:

Options:
(أ) probabilistic (ẓanniyya)
(ب) some are definitive and some are probabilistic
(ج) all of the above are incorrect
(د) definitive (qaṭʿiyya)

Response instructions:
- Respond with a single Arabic letter only, chosen from: أ, ب, ج, د
- Do not write any explanation or additional commentary
- Do not write any words other than the letter

## KNOW

### Arabic rendered prompt

أنت خبير متخصص في أصول الفقه، تتمتع بمعرفة عميقة ودقيقة في هذا المجال.

مهمتك هي الإجابة عن السؤال بدقة وإيجاز، مع الاستشهاد بالأدلة الشرعية عند الحاجة.

السؤال: عدد أقسام الحكم التكليفي الخمسة.

تعليمات الإجابة:
- أجب باللغة العربية فقط
- بإيجاز دون التطرق إلى معلومات غير مطلوبة
- إن كانت الإجابة تتضمن ذكر دليل شرعي، تحقق من الآية أو الحديث قبل كتابة الإجابة
- اكتب درجة ثقتك بإجابتك من 0% إلى 100%

**English translation**

You are a specialist in Principles of Jurisprudence with deep and precise expertise in this field.

Your task is to answer the question accurately and concisely, citing scriptural proofs (adilla sharʿiyya) when required.

Question: Enumerate the five categories of al-ḥukm al-taklīfī (the normative legal valuation of an act).

Response instructions:
- Answer in Arabic only
- Be concise and avoid including information not requested
- If the answer involves citing a scriptural proof, verify the verse or ḥadīth before writing it
- State your confidence in your answer from 0% to 100%

# COMP

## Arabic rendered prompt

.أنت خبير متخصص في أصول الفقه، تتمتع بمعرفة عميقة ودقيقة في هذا المجال

.مهمتك هي قراءة النص بعناية والإجابة عن السؤال بناءً على المعلومات الواردة فيه

:النص
:بيان أنواع مفهوم المخالفة، لأن هذا المفهوم يتنوع بحسب القيد الذي قيد به منطوق النص إلى خمسة أنواع :١- مفهوم الوصف :كقوله تعالى في بيان المحرمات "}وحلائل أبنائكم الذين من أصلابكم] {النساء :٢٣[، مفهوم المخالفة حلائل الأبناء الذين ليسوا من الأصلاب كابن الابن رضاعا، وكقول الرسول" :في السائمة زكاة مفهوم المخالفة المعلوفة التي ليست سائمة، وكقوله" :من باع نخلة مؤبرة فثمرتها للبائع "٢- مفهوم الغاية :كقوله تعالى} :فإن طلقها فلا تحل له من بعد حتى تنكح زوجا غيره] {البقرة :٢٣٠[، مفهوم المخالفة إذا تزوجت المطلقة ثلاثا زوجا غير مطلقها، وقوله تعالى} :وكلوا واشربوا حتى يتبين لكم الخيط الأبيض من الخيط الأسود من الفجر] {البقرة :١٨٧[، مفهوم المخالفة إذا تبين الأبيض من الأسود من الفجر .٣- مفهوم الشرط :كقوله تعالى} :وإن كن أولات حمل فأنفقوا عليهن] {الطلاق :٦[ مفهوم المخالفة إن كن لسن أولات حمل، وكقوله تعالى} :فإن طبن لكم عن شيء منه نفسا فكلوه هنيئا مريئا] {النساء :٤[، مفهوم المخالفة إذا لم تطب نفس الزوجة عن شيء من مهرها .٤- مفهوم العدد :كقوله تعالى} :فاجلدوهم ثمانين جلدة] {النور :٤[، مفهوم المخالفة الأقل والأكثر من ثمانين، وكقوله تعالى} :فمن لم يجد فصيام ثلاثة أيام] {البقرة :١٩٦[، مفهوم المخالفة الأقل والأكثر من ثلاثة .٥- مفهوم اللقب :كقوله تعالى} :محمد رسول الله] {الفتح :٢٩ [مفهوم المخالفة غير محمد، وكقول الرسول" :في البر صدقة :"مفهوم المخالفة غير البر، وكقوله تعالى} :حرمت عليكم أمهاتكم] {النساء :٢٣[، مفهوم المخالفة غير الأمهات

.السؤال :عدد الأنواع الخمسة لمفهوم المخالفة

:تعليمات الإجابة
أجب باللغة العربية فقط -
- بإيجاز دون التطرق إلى معلومات غير مطلوبة
استخدم النص كمصدر أساسي -
يمكنك إضافة معلومات عامة للمقارنة أو التحليل إذا طلب السؤال ذلك -
إن كانت الإجابة تتضمن ذكر دليل شرعي، تحقق من الآية أو الحديث قبل كتابة الإجابة -
%اكتب درجة ثقتك بإجابتك من 0 %إلى 100 -

## English translation

You are a specialist in Principles of Jurisprudence with deep and precise expertise in this field. Your task is to read the passage carefully and answer the question based on the information it contains.

Passage: An exposition of the types of mafhūm al-mukhālafa (implication by contrast). This category divides, according to the qualifier that restricts the wording of the text, into five types.

(1) Implication by description (mafhūm al-waṣf): e.g., God's statement specifying the prohibited categories of marriage, “and the wives of your sons who are from your loins” (Q 4:23) — the implication by contrast is the wives of sons not from one's loins, such as the son through fosterage; and the Prophet's saying, “Zakāt is due on free-grazing livestock,” whose implication by contrast is stall-fed livestock that does not graze freely; and his saying, “Whoever sells a pollinated date palm, the fruit is the seller's.”

(2) Implication by end-limit (mafhūm al-ghāya): e.g., “If he divorces her, she is not lawful for him thereafter until she marries another husband” (Q 2:230) — the implication by contrast is that she becomes lawful once she has married another husband after the triple divorce; and “and eat and drink until the white thread becomes distinguishable to you from the black thread of dawn” (Q 2:187) — the implication by contrast applies once the white thread can be distinguished from the black at dawn.

(3) Implication by condition (mafhūm al-shart): e.g., “And if they are pregnant, then spend on them” (Q 65:6) — the implication by contrast is that they are not pregnant; and “But if they remit any of it to you of their own accord, then enjoy it” (Q 4:4) — the implication by contrast is that the wife has not freely remitted any of her dower.

(4) Implication by number (mafhūm al-ʿadad): e.g., “Lash them eighty lashes” (Q 24:4) — the implication by contrast is anything less or more than eighty; and “Whoever does not find [the means], then fasting three days” (Q 2:196) — the implication by contrast is anything less or more than three.

(5) Implication by proper name/designation (mafhūm al-laqab): e.g., “Muḥammad is the Messenger of God” (Q 48:29) — the implication by contrast is anyone other than Muḥammad; and the Prophet's saying, “Charity is due on wheat,” whose implication by contrast is anything other than wheat; and “Forbidden to you are your mothers” (Q 4:23) — the implication by contrast is anyone other than the mothers.

Question: Enumerate the five types of mafhūm al-mukhālafa.

Instructions: Answer in Arabic only; be concise and avoid including information not requested; use the passage as the primary source; you may add general information for comparison or analysis only if the question asks for it; if the answer involves citing a scriptural proof, verify the verse or ḥadīth before writing it; state your confidence from 0% to 100%.

## S6. Judge prompts for open-ended scoring

This section reproduces the judge prompts used to score COMP and KNOW responses against gold reference answers. The same prompt templates are used for model and human responses. Judges return a scalar score in [0,1] at 0.1 increments and a short explanation. Gold answers may be single-answer, any-one-of, or any-K-of references (Table S5).

*Table S5. Rendering of gold-reference formats in the judge prompt.*

| Gold-reference type | Arabic rendering | English rendering |
|---|---|---|
| Single answer | — | The answer string, rendered directly. |
| Any-one-of (multiple candidates, K = 1) | أي من الأجوبة أدناه يعتبر صحيح. | Any of the answers below is acceptable, followed by a bulleted list of candidates. |
| Any-K-of (multiple candidates, K > 1) | ذكر أي K من الأجوبة أدناه يعتبر صحيح. | Providing any K of the answers below is acceptable, followed by a bulleted list of candidates. |

### KNOW judge prompt

#### Arabic prompt

اقرأ السؤال وافهم محتواه، ثم اطّلع على الإجابة الصحيحة بعناية.

بعد ذلك، قارن إجابة النموذج بالإجابة الصحيحة، مع تقييم مدى صلتها، دقتها، وشمولها.

السؤال:
{[نص السؤال]}

الإجابة الصحيحة:
{[الإجابة الصحيحة]}

إجابة النموذج:
{[إجابة النموذج]}

بعد قراءة كل المعلومات بعناية، قيّم إجابة النموذج وفق مقياس من 0 إلى 1 بناءً على:
- صحة الإجابة
- الدقة
- الشمولية
- وأن لا يستند التقييم على طول أو قصر إجابة النموذج.

معايير التقييم:
- 0 تعني أن الإجابة خاطئة تمامًا.
- 1 تعني أن الإجابة صحيحة وكاملة.

- يجب أن تكون الدرجة بين 0 و 1 وبخطوة 0.1.
- إذا كانت الإجابة ناقصة أو غير دقيقة، يتم منح درجة أقل بناءً على مدى صحة وشمولية الإجابة.
- إذا احتوت الإجابة على معلومات خاطئة، يتم خصم النقاط بناءً على خطورة الخطأ.

إخراج التقييم (بالصيغة المطلوبة):
يجب أن يكون الإخراج بصيغة JSON مع الحقول التالية:

```
{
  "score": [Score],
  "explanation": "[Explanation]"
}
```

حيث score هو الدرجة المحسوبة بين 0 و 1 (رقم)، و explanation هو شرح موجز يوضح سبب منح الدرجة (نص).

**English translation**

Read the question and understand its content, then review the correct answer carefully. Then compare the response to the correct answer, evaluating its relevance, accuracy, and completeness.

Question: [question text].

Correct answer: [correct answer].

Response: [model or human response].

After reading all information carefully, score the response on a 0–1 scale based on correctness, precision, and completeness. The length of the response must not influence the score.

Scoring criteria: 0 = completely incorrect; 1 = fully correct and complete; scores must be in {0.0, 0.1, ..., 1.0}; incomplete or imprecise answers receive a proportionally lower score; factual errors incur deductions proportional to their severity.

Required output (JSON):

```
{
 "score": [Score],
 "explanation": "[Explanation]"
}
```

where score is a number in [0,1] and explanation is a concise justification of the assigned score.

## COMP judge prompt

The COMP prompt is identical to the KNOW prompt, except that it includes the source passage before the question and instructs the judge to read the passage first. Only the differing portions are reproduced here.

**Modified Arabic opening**

اقرأ النص جيدًا وافهم محتواه، ثم اطّلع على السؤال والإجابة الصحيحة بعناية.

بعد ذلك، قارن إجابة النموذج بالإجابة الصحيحة، مع تقييم مدى صلتها، دقتها، وشمولها.

النص:
{[نص المقطع]}

السؤال:
{[نص السؤال]}

الإجابة الصحيحة:
{[الإجابة الصحيحة]}

إجابة النموذج:
{[إجابة النموذج]}

**English translation**

Read the passage carefully and understand its content, then review the question and the correct answer carefully. Then compare the response to the correct answer, evaluating its relevance, accuracy, and completeness.

Text: [passage text].

Question: [question text].

Correct answer: [correct answer].

Response: [model or human response].

The same 0–1 scoring criteria and JSON output format used for KNOW are then applied.

## S7. Judge model selection protocol

Five candidate judges were screened by a domain-trained MA-level Islamic Studies reviewer; the screening protocol was then reviewed with the senior professor panel. The screen used 20 held-out KNOW items, comprising 10 Intermediate and 10 Advanced scholarly-demand items. For each item, the reviewer inspected the source text, scholarly-demand tier, question, gold reference answer, model response, and every candidate judge's score and explanation in parallel. Candidates were assessed for discounting behavior, factual accuracy, coverage, and methodological validity (Table S6).

*Table S6. Candidate judge models and their role in the final evaluation pipeline. The two production judges were chosen for complementary scoring tendencies; the head judge was reserved for high-disagreement cases.*

| Candidate | Final role | Expert-review rationale |
| --- | --- | --- |
| GPT-4o | -- | Similar to GPT-5.2 in analytical depth but slightly more lenient on incomplete answers; not retained. |
| GPT-5.2 | Production judge | Strong analytical depth and partial-credit reasoning; somewhat tolerant of textual mismatch. |
| Gemini-2.5-Flash | Production judge | Stricter on textual precision and source adherence; complements GPT-5.2's analytical leniency. |
| DeepSeek-Chat | -- | More lenient on incomplete answers and less sensitive to discipline-specific distinctions in the screen. |
| gemini-3.1-pro-preview | Head judge | Best matched the expert's judgments on the held-out Arabic scholarly items; reserved for high-disagreement adjudication ($\lvert j1 - j2 \rvert > 0.2$). |